\PassOptionsToPackage{table}{xcolor}
\documentclass{bmvc2k}
\usepackage{tabularx}
\usepackage{bm}
\usepackage{booktabs}
\usepackage{multirow}
\usepackage{pifont}
\usepackage{float}
\newcommand{\cmark}{\ding{51}}
\newcommand{\xmark}{\ding{55}}
\usepackage{soul}
\usepackage{enumitem}
\usepackage{wrapfig}
\usepackage{booktabs}
\definecolor{r}{rgb}{1,0.8,0.8}
\definecolor{y}{rgb}{1,1,0.8}

\title{RUSplatting: Robust 3D Gaussian Splatting for Sparse-View Underwater Scene Reconstruction}

\addauthor{Zhuodong	Jiang}{davidjiang326@gmail.com}{1}
\addauthor{Haoran Wang}{yp22378@bristol.ac.uk}{1}
\addauthor{Guoxi Huang}{guoxi.huang@bristol.ac.uk}{1}
\addauthor{Brett Seymour}{Brett_Seymour@nps.gov}{2}
\addauthor{Nantheera Anantrasirichai}{n.anantrasirichai@bristol.ac.uk}{1}

\addinstitution{
 Visual Information Laboratory, \\
 University of Bristol, \\ Bristol, UK
}
\addinstitution{
 Submerged Resources Center, \\
 National Park Service, \\
 Denver, CO, USA
}

\runninghead{Jiang, Wang, Huang~\etal}{RUSplatting for Sparse-View Underwater 3D}

\def\etal{\emph{et al}\bmvaOneDot}

\begin{document}

\maketitle

\begin{abstract}
Reconstructing high-fidelity underwater scenes remains a challenging task due to light absorption, scattering, and limited visibility inherent in aquatic environments. This paper presents an enhanced Gaussian Splatting-based framework that improves both the visual quality and geometric accuracy of deep underwater rendering. We propose decoupled learning for RGB channels, guided by the physics of underwater attenuation, to enable more accurate colour restoration. To address sparse-view limitations and improve view consistency, we introduce a frame interpolation strategy with a novel adaptive weighting scheme. Additionally, we introduce a new loss function aimed at reducing noise while preserving edges, which is essential for deep-sea content. We also release a newly collected dataset, Submerged3D, captured specifically in deep-sea environments. Experimental results demonstrate that our framework consistently outperforms state-of-the-art methods with PSNR gains up to 1.90dB, delivering superior perceptual quality and robustness, and offering promising directions for marine robotics and underwater visual analytics. The code of RUSplatting is available at \url{https://github.com/theflash987/RUSplatting} and the dataset Submerged3D can be downloaded at \url{https://zenodo.org/records/15482420}.
\end{abstract}

\section{Introduction}
\label{sec:intro}
Underwater exploration has a wide range of applications, including marine ecosystem research, underwater archaeology, and geology. These studies heavily rely on effective data acquisition and visualisation. However, underwater exploration is significantly constrained by the challenging environment, including the risks associated with diving, a shortage of professional divers, and the high costs of both equipment and operations. Additionally, certain viewpoints or data may lack sufficient overlap, making it difficult to produce high-quality 3D reconstructions. Consequently, there is a pressing need for technologies capable of effectively reconstructing complex marine environments using limited data, enabling onshore visualisation and subsequent analysis.

Advancements in Novel View Synthesis (NVS) have introduced new possibilities for reconstructing and visualising environments in 3D space from multi-view image data. These methods allow the generation of unseen viewpoints using limited training images. A notable breakthrough in NVS is Neural Radiance Fields (NeRF)~\cite{mildenhall2021nerf}, which implicitly models 3D scenes. Despite their impressive performance, NeRFs require substantial training time and are impractical for real-time or interactive applications~\cite{chen2024survey, wu2024recent}. Moreover, their representations are inherently difficult to interpret or modify intuitively~\cite{wu2024recent}, limiting their flexibility in dynamic environments, such as underwater. In contrast, 3D Gaussian Splatting (3DGS)~\cite{kerbl20233d} reconstructs scenes explicitly using Gaussian point sets, offering faster training, higher rendering speeds and refresh rates, as well as more flexible scene manipulation. 

Nevertheless, most existing NeRF-based and 3DGS-based methods focus on reconstructing scenes in clear media and therefore struggle to produce high-quality results in hazy and distorted environments like underwater. Due to the physical properties of water, underwater imaging is affected by attenuation and scattering, where the former causes colour casting and the latter results in low contrast~\cite{liu2024aquatic}. It is therefore essential to account for these properties when modelling the entire scene.

In the face of these challenges, we propose a new framework: \textbf{R}obust \textbf{U}nderwater scene representation using 3D Gaussian \textbf{Splatting} (RUSplatting), which demonstrates greater robustness than previous works, such as UW-GS~\cite{wang2024uw} and WaterSplatting~\cite{li2024watersplatting}. Previous approaches integrate medium effects via affine transformations of Gaussian attributes but assume dense, temporally coherent input frames, limiting effectiveness under sparse-view conditions. They also do not address distortions common in deep-sea imaging, such as strong noise and severe colour imbalance.

\begin{figure*}
  \begin{center}
  \includegraphics[width=\linewidth]{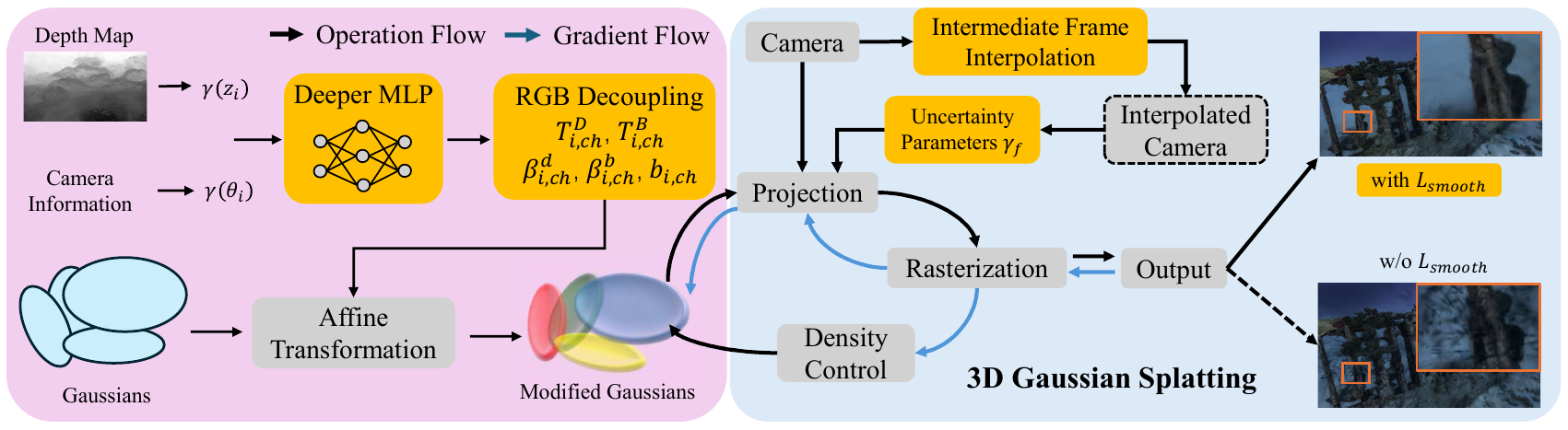}
  \end{center} \vspace{-5mm}
  \caption{Pipeline of RUSplatting. Yellow highlights indicate the proposed contributions: (1) colour-channel decoupling with a deeper MLP for improved estimation of underwater medium parameters; (2) IFI to enrich representations and enhance frame overlap; (3) an AFW strategy to suppress contributions from poorly interpolated frames; and (4) an ESL to reduce noise while preserving fine structural details.}
  \label{overview}
\end{figure*}

Our proposed RUSplatting framework, shown in Fig.~\ref{overview}, further advances underwater 3D reconstruction by explicitly decoupling the estimation of attenuation and scattering parameters for individual RGB channels instead of treating them uniformly, enabling more accurate and realistic colour restoration. As discussed in~\cite{peng2017underwater}, the light attenuation underwater has different degrees of changes due to the wavelength. Red light, which has a longer wavelength, is absorbed more rapidly than green and blue light. Furthermore, to mitigate image noise prevalent in deep-sea conditions, we introduce an Edge-Aware Smoothness Loss (ESL) to directly optimise the reconstruction pipeline, dynamically reducing noise while preserving critical geometric edges. Additionally, to address the inherent challenges of sparse-view datasets, we incorporate an Intermediate Frame Interpolation (IFI) mechanism, enriching viewpoint coverage and increasing initialised Gaussian points. However, since interpolated frames may contain artefacts that could deteriorate the learning process, we propose an Adaptive Frame Weighting (AFW) strategy that adjusts the loss based on the predictive uncertainty~\cite{kendall2017uncertainties, kendall2018multi}. Finally, we introduce a new dataset, Submerged3D, capturing detailed objects in deep-sea environments, supporting future research in underwater robotics, archaeology, and marine biology.

In summary, our key contributions are as follows:
\begin{itemize}[leftmargin=15pt, nolistsep]
    \item We propose a novel framework, RUSplatting, for 3D representation of deep underwater scenes under sparse-view conditions.
    \item RUSplatting enables more accurate colour estimations through colour-channel decoupling, which particularly improves clarity in high-turbidity conditions.
    \item We introduce a novel Edge-Aware Smoothness Loss (ESL) to suppress noise commonly encountered in deep-sea, low-light conditions while preserving structural details.
    \item We introduce Intermediate Frame Interpolation (IFI) and Adaptive Frame Weighting (AFW) to address viewpoint sparsity.
    \item We present Submerged3D, the first dataset captured in deep underwater environments, highlighting real-world challenges such as limited lighting and severe turbidity.
\end{itemize}

\noindent Extensive experiments on both existing and new datasets demonstrate that RUSplatting consistently outperforms state-of-the-art methods across various evaluation metrics, with PSNR improvements of up to 21.37\%, 4.99\% and 5.09\% compared to SeeThru-NeRF~\cite{levy2023seathru}, WaterSplatting~\cite{li2024watersplatting} and UW-GS~\cite{wang2024uw}, respectively.

\section{Related Work}

\paragraph{Underwater images.}
Processing underwater images remains a major challenge in underwater computer vision due to colour distortion and noise arising from water's physical properties and its impact on light propagation~\cite{gonzalez2023survey}. The Underwater Image Formation Model (UIFM)~\cite{chiang2011underwater,drews2016underwater} was introduced to describe this process, accounting for light absorption and backscattering. While the original UIFM assumes a homogeneous water column, the Revised UIFM (R-UIFM)~\cite{akkaynak2018revised} improves realism by allowing spatial variability in water properties, making it widely adopted in underwater image processing. For example, PRWNet~\cite{huo2021efficient} uses a deep learning-based multi-stage refinement approach that enhances image quality by addressing spatial and frequency domain degradations through Wavelet Boost Learning. However, the survey in~\cite{huang2024visual} shows that applying enhancement prior to 3D reconstruction results in poorer performance compared to embedding the physical properties of water directly into the reconstruction process.

\paragraph{NVS methods for underwater.}
While the NeRF-based and GS-based technologies have rapidly advanced, most struggle in scattering media such as fog or water. To tackle this limitation, ScatterNeRF~\cite{ramazzina2023scatternerf} was introduced to effectively separate objects from the scattering medium in foggy environments. SeaThru-NeRF~\cite{levy2023seathru} extends this line of research by estimating water medium coefficients using a novel architecture grounded in the Revised Underwater Image Formation Model (R-UIFM)~\cite{akkaynak2018revised}. AquaNeRF~\cite{gough2025aquanerf} improves SeaThru-NeRF with single surface rendering to avoid distractors. Tang \etal.~\cite{tang2024neural} propose a neural representation that incorporates dynamic rendering and tone mapping to enable more accurate reconstruction of underwater scenes under varying illumination conditions. 

For 3DGS-based methods, UW-GS~\cite{wang2024uw} and WaterSplatting~\cite{li2024watersplatting} employ MLP-based colour appearance modules to address colour casting. In contrast, SeaSplat~\cite{yang2025seasplat} optimises backscatter parameters using a dark channel prior loss. UW-GS~\cite{wang2024uw}, UDR-GS~\cite{du2024udrgs}, and Aquatic-GS~\cite{liu2024aquatic} leverage DepthAnything~\cite{yang2024depth} to estimate depth from 2D inputs and incorporate it into their loss functions. UW-GS~\cite{wang2024uw} further improves reconstruction of distant objects through a physics-driven density control module.

\section{Preliminaries}

The 3DGS~\cite{kerbl20233d} employs a collection of spatial Gaussian cloud points to represent a 3D scene. To synthesise a novel view, these Gaussians are projected onto a 2D image plane. The projected Gaussian is defined as
\begin{equation}
G_i(\bm{x}) = \exp\left( -\frac{1}{2} (\bm{x} - \bm{\mu}_i)^\top \Sigma_i^{-1} (\bm{x} - \bm{\mu}_i) \right),
\end{equation}
where the center of Gaussian $G_i$ is located at $\bm{\mu}_i$, with covariance matrix $\Sigma_i$. Both $\bm{\mu}_i$ and $\bm{x}$ lie on the 2D image plane. Each Gaussian $i$ is further characterised by an opacity $\alpha_i$ and a view-dependent colour appearance $c_i$, parameterised using spherical harmonics (SH). The final pixel colour $C$ is computed via $\alpha$-blending as follows:
\begin{equation}
\label{colour}
C = \sum_{i=1}^{N} \alpha_i c_i(\mathbf{v}) \prod_{j=1}^{i-1}(1 - \alpha_j), \quad \text{where} \ c_i(\mathbf{v}) = \sum_{l=0}^{L} \sum_{m=-l}^{l} c_{i,lm} \cdot Y_{lm}(\mathbf{v}).
\end{equation}
Here, $c_{i,lm}$ are the learned SH coefficients for each RGB channel, and $Y_{lm}(\mathbf{v})$ are the real SH basis functions in the viewing direction $\mathbf{v}$, where $l$ and $m$ denote the degree and order, respectively. The $\alpha$-blending operation accounts for occlusions among Gaussians, resulting in a more accurate and detailed final colour composition.

\paragraph{Underwater image formation.}
Compared to clear, onshore scenes, where scattering and absorption caused by the medium (i.e., air) can often be neglected, underwater scenes are heavily affected by these parameters, since they significantly degrade reconstruction quality~\cite{tang2024neural}. Therefore, it is crucial to consider these medium parameters during reconstruction. As modelled in~\cite{akkaynak2018revised}, the intensity of the underwater image is defined as follows:
\begin{equation}
    I_c = J \cdot T^D + B^{\infty} \cdot (1 - T^B),
    \label{uwimage}
\end{equation}
where $I_c$ denotes the colour captured by the camera, and $J$ represents the true object colour, free from any medium-induced effects. $B^{\infty}$ refers to the background light from an infinite distance. $T^D$ and $T^B$ are transmission terms representing the attenuation of direct and backscattered light, respectively. The $T^D$ and $T^B$ are defined as,
 \begin{equation}
T^D = \exp(-\beta^{d} \cdot z) \quad \text{and} \quad T^B = \exp(-\beta^{b} \cdot z),
   \label{eq:attnbs}
 \end{equation}
Here $\beta^{d}$ and $\beta^{b}$ are two colour medium coefficients, and $z$ is the line-of-sight distance from the camera to the 3D scene point rather than the water depth.

\section{Methodology}
The detailed structure of the proposed framework is presented in Fig.~\ref{overview}. The model estimates underwater medium parameters using an MLP with encoded depth and viewpoint position information as inputs. The MLP $f_\theta(\gamma_z(z), \gamma_\theta(\mathbf v))$ predicts, for each Gaussian and channel, the parameters $\{{T^D, T^B, \beta^d, \beta^b, b}\}$, which are jointly optimised under the main loss function described in Section.~\ref{loss}. Accordingly, the colours of the spatial 3D Gaussians are modified through the affine transformation. Then, the modified Gaussians, including additional Gaussians introduced by the frame interpolation process with learnable uncertainty weights, are projected onto the 2D plane. During training, an edge-aware smoothness loss is integrated into the model as a part of the loss function to remove noise from underwater images, and in the meantime maintaining geometric details.

\subsection{Colour-Channel Decoupling}

As discussed in numerous underwater studies~\cite{liu2024aquatic,li2025zerotigtemporalconsistency,huang2024visual}, the attenuation rates differ across red, green, and blue light channels. Therefore, the medium parameters for the RGB channels cannot be treated identically. To overcome this and achieve more accurate colour reconstruction, we decouple the channel-wise learning of the underwater medium. Consequently, the attenuation ($T^D$) and backscatter ($T^B$) factors for each colour channel $ch \in \{r, g, b\}$ are defined as:
\begin{equation}
T_{ch}^D = \exp(-\beta_{ch}^d \cdot z), \quad T_{ch}^B = \exp(-\beta_{ch}^b \cdot z).
\label{medium}
\end{equation}
Previous works use a 2-layer MLP~\cite{lee2024compact} to estimate the complex underwater medium parameters, however, it may fall short in capturing such complicated patterns as the study indicated in~\cite{he2016deep}. Thus, we employ a deeper MLP with 5 layers, selected based on a hyperparameter sensitivity analysis (see supplementary material for details). The final corrected colour of Gaussian $i$, $c_{i,ch}^m(v)$, is obtained through an affine transformation, as presented below, where $c_{i,ch}(v)$ is the observed view-dependent colour.
\begin{equation}
    c_{i,ch}^m(v) = T_{i,ch}^D \cdot c_{i,ch}(v) + (1 - T_{i,ch}^B) \cdot b_{i,ch}.
    \label{eq:cfinal}
\end{equation}
Here, $b_{i,ch}$ denotes a bias term specific to each Gaussian and channel, modelling the background light. It is assumed view-independent within the views observing Gaussian $i$, while view-dependent effects are captured by $c_{i,ch}(v)$. As Eq.~\ref{eq:cfinal} follows directly from Eq.~\ref{uwimage}, the observed channel-wise colour equals a scaled direct term plus an additive background light term, hence the mapping is affine.

\subsection{Intermediate Frame Interpolation (IFI)}
\label{ssec:frameinterp}
To address the limitations introduced by sparse views, we adopt a frame interpolation technique to synthesise intermediate frames between adjacent input images, leveraging them in two key aspects: i) enhancing the initial point cloud and ii) improving the RUSplatting optimisation, where we propose Adaptive Frame Weighting (AFW). Although various methods are available and should not significantly affect the final 3DGS results, we use Real-time Intermediate Flow Estimation (RIFE)~\cite{huang2022real}, a state-of-the-art deep learning-based approach that employs bidirectional optical flow and knowledge distillation to enable real-time processing. 

\paragraph{Initial point cloud.} The initial point cloud (obtained using COLMAP~\cite{schoenberger2016sfm}) acts as a rough geometric framework for the final model. Its quality and density directly influence the accuracy of the rendering. By incorporating interpolated frames and re-running COLMAP on the augmented dataset, additional initialisation points are introduced, enabling a denser distribution of Gaussians and improving the overall reconstruction precision and detail. Without IFI, the model accesses an average of 17,635 initialisation points across all datasets. After adopting IFI, this number increases significantly by 55.7\% to an average of 27,461 points. Detailed results for each dataset are provided in the supplementary material.

\paragraph{Adaptive Frame Weighting (AFW) $\gamma$.} In sparse view scenarios, adjacent frames are significantly spaced apart, making frame interpolation challenging. The large spatial shifts between frames may introduce artefacts in certain geometric structures of the interpolated frames, which may degrade the quality of 3D reconstruction during optimisation when these frames are used. Inspired by~\cite{kendall2017uncertainties, kendall2018multi}, we propose an AFW mechanism, introducing an uncertainty learnable parameter $\gamma$  for each interpolated frame $f$ that is optimised along with the training. Accordingly, the loss function for the interpolated frame $L_f$ is updated as follows.
\begin{equation}
\label{uncertain}
    {L}'_f = \frac{1}{2} \cdot \gamma_f \cdot L_{f} - \frac{1}{2} \cdot \alpha \cdot \log(\gamma_f),
\end{equation}
where $\alpha$ controls the relative weight of the uncertainty regularisation term. $\alpha$ is a hyperparameter selected within the range $[0, 1]$. Adapting such weights ensures that interpolated frames aid the training while alleviating the effect caused by the potential artefacts.

\subsection{Edge-Aware Smoothness Loss (ESL)}
\label{smooth_loss}
Inspired by bilateral filtering, which smooths noisy images while preserving edges, we propose a novel loss function that behaves dually, removing noise and preserving geometric edges. A key advantage of our approach is that the functionality is embedded into the end-to-end training framework without relying on multiple predefined parameters like traditional bilateral filters. Instead, we utilise the underlying 3D geometry to guide edge-preserving smoothness in the rendered images.

Our edge-aware smoothness loss is defined as,
\begin{equation}
\label{smooth_loss_eq}
L_\text{Smooth}(I, D) = \sum_{i,j} \left( w^x_{i,j} \cdot \left| I_{i,j} - I_{i,j+1} \right| + w^y_{i,j} \cdot \left| I_{i,j} - I_{i+1,j} \right| \right),
\end{equation}
where $(i,j)$ indexes the spatial coordinates in the image plane, $I$ denotes the rendered image and the $D$ denotes the pseudo depth map generated from the Depth-Anything-V2~\cite{yang2024depthv2}. The edge-aware weights $w^x_{i,j}$ and $w^y_{i,j}$, corresponding to the horizontal and vertical directions, respectively, are computed using depth discontinuities,
\begin{equation}
w^x_{i,j} = \exp\left(-\lambda_b \cdot \left| D_{i,j} - D_{i,j+1} \right| \right), \quad
w^y_{i,j} = \exp\left(-\lambda_b \cdot \left| D_{i,j} - D_{i+1,j} \right| \right).
\end{equation}
Here, $\lambda_b$ controls the edge sensitivity and is selected via grid search in the range $[0, 5]$.

\subsection{Loss Function}
\label{loss}
The entire model is trained end-to-end using a combination of four loss terms, including three common losses (described below) and our proposed edge-aware smoothness loss, $L_\text{Smooth}$ (described in Section~\ref{smooth_loss}), with a weight $\lambda_s$ of 0.2. Additionally, we integrate the adaptive frame weighting for interpolated frames as described in Section~\ref{ssec:frameinterp}. Consequently, the final loss function is in the form of,
\begin{equation}
    L_{final} =  L_{\text{Rec}} + L_{\text{Depth}} + L_{\text{g}} + \lambda_s L_\text{Smooth},
\end{equation}
for the original frame, while for the interpolated frame, the loss function turns to,
\begin{equation}
    L_{final}' =  \frac{1}{2} \cdot \gamma \cdot L_{final} - \frac{1}{2}\cdot \alpha \cdot \log(\gamma).
\end{equation}

\paragraph{Reconstruction Loss $L_\text{Rec}$.} 
To evaluate the differences between the reconstructed image and the ground truth image, the $L_\text{Rec}$ is defined as
$
    L_{\text{Rec}} = \lambda_{r} \cdot \omega \cdot L_1 + (1 - \lambda_r) \cdot \omega \cdot L_{\text{SSIM}}
$, 
where the $\lambda_r$ is a loss weight to balance the $L_1$ and $L_{D-SSIM}$. $\omega$ denotes the Binary Motion Mask (BMM)~\cite{wang2024uw} that remove the distractor and marine snow artefacts. $L_1$ refers to the absolute error loss and $ L_{\text{SSIM}}$ is the SSIM loss. 

\paragraph{Depth-supervised Loss $L_{\text{Depth}}$.} 
The depth information plays a vital role as the depth $z$ information is involved in the estimation of the underwater medium parameters, as illustrated in Eq.~\ref{medium}. Thus, the $L_{\text{Depth}}$ is included and defined as,
$
    L_{\text{Depth}} = \lambda_d \cdot \| \hat{D} - D \| + \lambda_{ca} \cdot ( \sum_{ch \in \{r,g,b\}} \sum_{n \in \{d,b\}} \|z^{n}_{ch} - D\| )
$, 
 where $\lambda_d$ and $\lambda_{ca}$ are loss weights, $D$ is the pseudo depth map, obtained from Depth-Anything-V2~\cite{yang2024depthv2}, and $\hat{D}$ is the rendered depth map from the RUSplatting. $z_{ch}^n$ denotes the inferred depth under direct ($d$) and backscatter ($b$) lighting conditions for each colour channel $ch$.

\paragraph{Grey-world Assumption Loss $L_g$.} Following~\cite{fu2022unsupervised}, $L_g$ is built based on the assumption that in a natural scene, the average value of each colour channel of the image should be equal. With this restriction, it can promote the colour balance and aid in the accurate estimation of the medium parameters. $L_g = \sum_{ch \in \{r, g, b\}} \left\| \mu(J_{ch}) - 0.5 \right\|_2^2$.
Here, $J_{ch}$ denotes the restored pixel values for all three colour channels, ranging between 0 to 1. $\mu(\cdot)$ denotes the mean value of a channel. It is worth noting that the $0.5$ here corresponds to the ideal mean intensity for the colour channel under the grey-world assumption.

\section{Experiments}
\begin{table}
    \begin{center}
    \scriptsize
    \renewcommand{\arraystretch}{1.2}
    \setlength{\tabcolsep}{3pt}
    \resizebox{\textwidth}{!}{

    \begin{tabular}{|rl|ccc|ccc|ccc|}
    \hline
    \multicolumn{2}{c|}{Dataset} & \multicolumn{3}{c|}{SeaThru-NeRF} & \multicolumn{3}{c|}{S-UW} & \multicolumn{3}{c|}{Submerged3D} \\
    \cmidrule(lr){3-5} \cmidrule(lr){6-8} \cmidrule(lr){9-11}
    & Method & PSNR $\uparrow$ & SSIM $\uparrow$ & LPIPS $\downarrow$ & PSNR $\uparrow$ & SSIM $\uparrow$ & LPIPS $\downarrow$ & PSNR $\uparrow$ & SSIM $\uparrow$ & LPIPS $\downarrow$ \\
    \hline
    & Instant-NGP~\cite{muller2022instant} & 23.7944 & 0.6317 & 0.4434 & 18.9001 & 0.4693 & 0.3966 & 19.0266 & 0.5107 & 0.4977 \\
    & SeaThru-NeRF~\cite{levy2023seathru} & 26.0604 & 0.7806 & 0.3398 & 21.2253 & 0.5948 & 0.4104 & 19.1967 & 0.4967 & 0.5685 \\
    & 3DGS~\cite{kerbl20233d} & 26.8353 & 0.9011 & 0.1826 & 23.7563 & 0.7745 & 0.2231 & 22.6551 & 0.7544 & 0.3288 \\
    & WaterSplatting~\cite{li2024watersplatting} & 27.3924 & 0.8475 & 0.1870 & \cellcolor{y}24.8788 & \cellcolor{y}0.7915 & 0.2159 & \cellcolor{y}24.5832 & 0.7517 & 0.3172 \\
    & UW-GS~\cite{wang2024uw} & \cellcolor{y}28.2557 & \cellcolor{y}0.9105 & \cellcolor{y}0.1750 & 24.1833 & 0.7838 & \cellcolor{y}0.2050 & 24.3409 & \cellcolor{y}0.7723 & \cellcolor{y}0.3107 \\
    & RUSplatting (ours) & \cellcolor{r}\textbf{29.2928} & \cellcolor{r}\textbf{0.9163} & \cellcolor{r}\textbf{0.1665} & \cellcolor{r}\textbf{25.5955} & \cellcolor{r}\textbf{0.8138} & \cellcolor{r}\textbf{0.1964} & \cellcolor{r}\textbf{25.7990} & \cellcolor{r}\textbf{0.7724} & \cellcolor{r}\textbf{0.3008} \\
    \hline 
\end{tabular}

    }
    \end{center}
    \caption{Performance comparison between RUSplatting and five baseline models on three datasets. The results are the average values over all four scenes in each dataset. $\uparrow$ indicates that higher values are better, while $\downarrow$ indicates that lower values are better. Red colour with bold denotes the best result and yellow denotes the second best.}
    \label{underwater_comparison}
\end{table}
\subsection{Implementation details}
\paragraph{Datasets.}
\begin{figure*}
  \begin{center}
  \includegraphics[width=\linewidth]{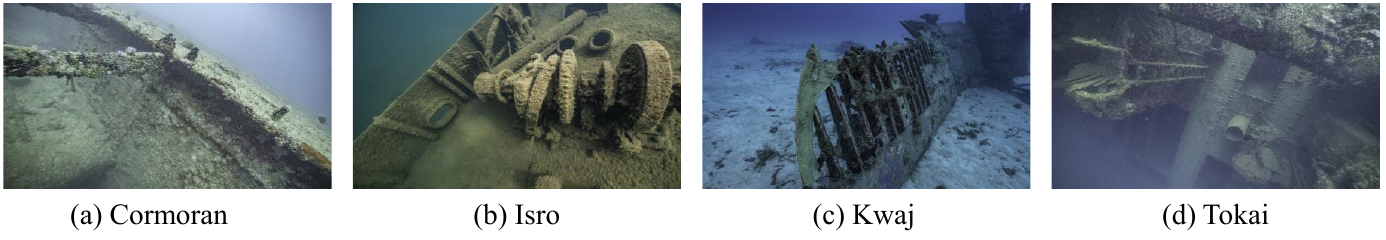}
  \end{center} \vspace{-5mm}
  \caption{Sample images from Submerged3D dataset.}
  \label{samples}
\end{figure*}
Three datasets were used: SeaThru-NeRF~\cite{levy2023seathru}, S-UW~\cite{wang2024uw} and our newly collected dataset, Submerged3D. Submerged3D consists of four scenes, each containing 20 RGB 720p images focusing on shipwrecks captured in the deep sea. It is challenging for 3D reconstruction due to sparse viewpoints and low-light environment conditions. Examples of the scenes in Submerged3D are shown in Fig.~\ref{samples}. To create the training and test splits, we assign every 8th image in the dataset to the test set and use the rest for training. SeaThru-NeRF, S-UW, and our Submerged3D provide no ground-truth depth, therefore we use Depth-Anything-V2~\cite{yang2024depthv2} pseudo-depth as weak supervision.

\paragraph{Experimental Setup.}
We utilised COLMAP~\cite{schoenberger2016sfm} to initialise the point cloud and estimate the camera poses for the image sequences. For each pair of adjacent images, only one interpolated frame is generated. The experiments are conducted on a single RTX 4090 GPU. All experiments are run for 20,000 iterations. $\lambda_r$, $\lambda_d$, and $\lambda_{ca}$ are set to 0.8, 0.1, and 1, respectively.

\subsection{Evaluation} We compared our results with the Instant-NGP~\cite{muller2022instant}, SeaThru-NeRF~\cite{levy2023seathru}, 3DGS~\cite{kerbl20233d}, WaterSplatting~\cite{li2024watersplatting} and UW-GS~\cite{wang2024uw}. For a fair comparison, we used their official implementations and trained each method on the same image sequences using our dataset split strategy. The quality of the synthesised views was evaluated using PSNR, SSIM, and LPIPS metrics.


\paragraph{Quantitative Comparisons.} 
Tab.~\ref{underwater_comparison} presents comparisons with prior state-of-the-art methods, reporting the average rendering quality across all scenes in each dataset. We can observe that our RUSplatting consistently outperforms all baseline methods over three datasets across all metrics. Specifically, RUSplatting achieves PSNR improvements of 2.83, 3.01, and 3.84 dB over baseline methods on SeaThru-NeRF, S-UW and Submerged3D, respectively. It also yields SSIM gains of 12.53\%, 19.19\%, and 17.53\%, and reduces LPIPS by 37.29\%, 32.32\%, and 25.66\% on the same datasets.

\begin{figure}
  \begin{center}
  \includegraphics[width=\linewidth]{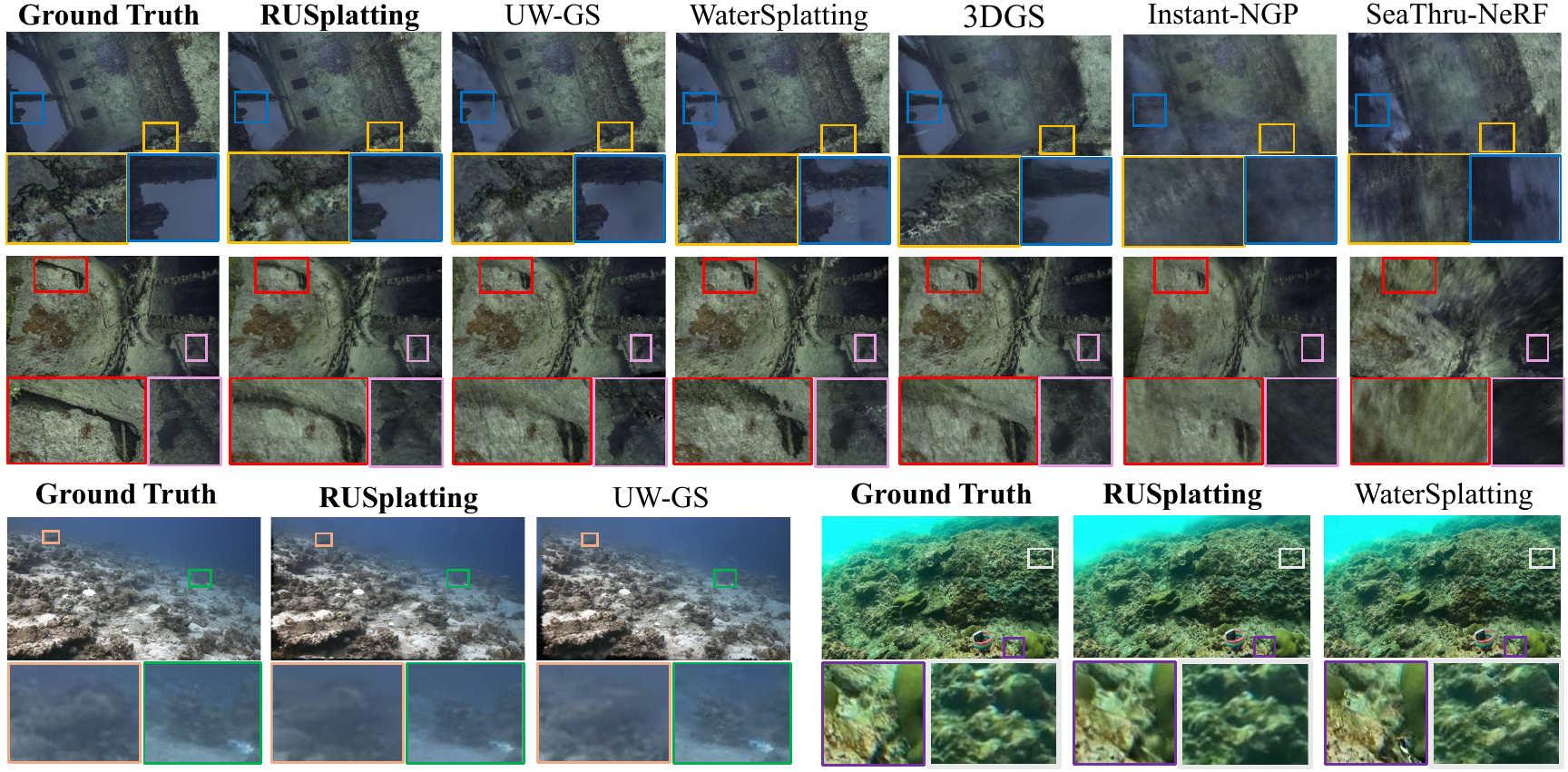}
  \end{center} \vspace{-4mm}
  \caption{Novel view rendering comparison on the Submerged3D, SeaThru-NeRF and S-UW datasets. The first two rows show results from the Cormoran, and Tokai scenes from Submerged3D, respectively. The left side of the third row shows the result from IUI-Reasea (SeaThru-NeRF), while the right side shows the Seabed from S-UW}
  \label{overall_results}
\end{figure}

\paragraph{Qualitative Comparisons.} The visual results shown in Fig.~\ref{overall_results} further demonstrate the superior performance of our method. Specifically, in the first row, all other models fail to produce a clear and accurate image and contain artefacts, as illustrated by the yellow and blue-outlined regions, whereas our method preserves fine details. Additionally, RUSplatting accurately reconstructs the gully structure in the second row, highlighted in red, which the other models fail to capture. Moreover, according to the third row, only our model renders high-quality results for both datasets. In contrast, the rendered images by UW-GS contain blurry artefacts (as shown in the green outlined region) and fail to capture the geometric structure in distant as displayed in the orange outlined area. Moreover, the outputs from WaterSplatting are noisy, as highlighted in the purple and grey outlined regions.

\subsection{Ablation Study}

\begin{figure}
\centering

\begin{minipage}[t]{0.35\textwidth} \vspace{8mm}
\vspace{0pt}
\centering
\scriptsize
\renewcommand{\arraystretch}{1.2}
\setlength{\tabcolsep}{3pt}
\begin{tabular}{|l|cccccc|}
    \hline
    Modules & M1 & M2 & M3 & M4 & M5 & RUSplatting \\
    \hline \hline
    Deeper MLP       & \xmark & \cmark & \cmark & \cmark & \cmark & \cmark \\
    RGB Decoupling   & \cmark & \xmark & \cmark & \cmark & \cmark & \cmark \\
    ESL             & \cmark & \cmark & \xmark & \cmark & \cmark & \cmark \\
    IFI w/o AFW      & \cmark & \cmark & \cmark & \xmark & \cmark & \cmark \\
    IFI w/ AFW       & \xmark & \xmark & \xmark & \xmark & \xmark & \cmark \\
    \hline
\end{tabular}
\end{minipage}
\hfill
\begin{minipage}[t]{0.5\textwidth}
\vspace{0pt}
\centering
\includegraphics[width=\linewidth]{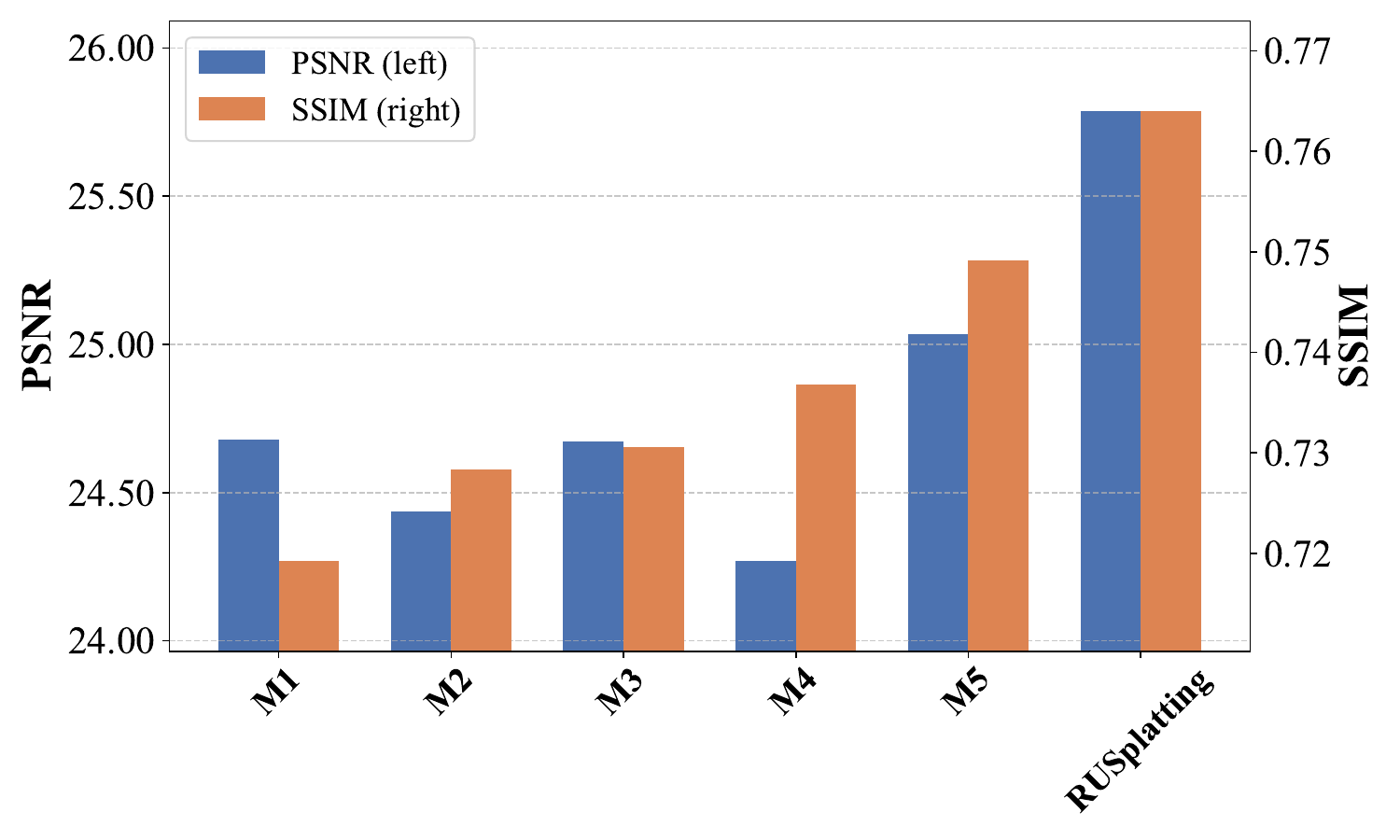}
\end{minipage}

\caption{
Left: Ablation study configurations. Right: Average results of all scenes. We construct five model variants (i.e., M1–M5) and compare them to RUSplatting.}
\label{fig:ablation}
\end{figure}

To evaluate the impact of each component, we conducted ablation studies, with configurations and results presented in Fig.~\ref{fig:ablation}. According to the right subfigure in Fig.~\ref{fig:ablation}, the following conclusions can be drawn.

\noindent \paragraph{Deeper MLP.}
The comparison between M1 and M5 demonstrates the effectiveness of a deeper architecture, as the SSIM increases significantly from 0.7192 to 0.7491, and PSNR rises notably from 24.68 to 25.03.

\noindent \paragraph{RGB Decoupling.}
We observe a substantial gain in SSIM (from 0.7284 to 0.7491) and PSNR (from 24.44 to 25.03) through the evaluation between M2 and M5. This indicates that RGB decoupling positively contributes to the overall performance.

\noindent \paragraph{ESL.}
Comparing M3  and M5, there is a notable improvement in SSIM from 0.7306 to 0.7491, and a substantial increase in PSNR from 24.67 to 25.03. This clearly highlights ESL’s essential role in preserving structural fidelity.

\noindent \paragraph{IFI.}
A 1.84 dB decrease from M5 to M4 regarding the PSNR, significantly evidences the importance of the IFI in terms of achieving high-quality pixel-wise reconstruction. Also, the SSIM drops from 0.7491 to 0.7368, indicating that the IFI is also critical in capturing sophisticated structure. It facilitates the model in obtaining more initialisation points and alleviates the impact of sparse perspectives.

\noindent \paragraph{AFW.}
Employing AFW (the full model, RUSplatting) results in the highest performance, achieving an SSIM of 0.7640 and a PSNR of 25.7874 on average across all datasets. By adjusting the uncertainty parameters of each interpolated frame, the model can capture more scene information for training by increasing the weight of high-quality interpolated frames, while reducing the weight of low-quality interpolated frames to reduce their potential impact on training. The improvement demonstrates the value of such adaptive weighting features for optimal image reconstruction quality.
\section{Conclusion}
This paper presents a new GS-based framework for deep underwater scene reconstruction. Specifically, we propose decoupled learning for the RGB channels, leveraging wavelength-dependent attenuation and scattering properties of underwater light. This approach enables more accurate colour reconstruction. To mitigate challenges posed by sparse viewpoints, we incorporate a frame interpolation mechanism that facilitates denser and more robust scene capture, along with adaptive frame weighting during training. Finally, we propose a new loss function that dynamically adjusts to image characteristics, effectively suppressing noise while preserving geometric structure. Collectively, our RUSplatting framework significantly enhances the visual fidelity, structural consistency, and colour accuracy of underwater reconstructions. However, as RIFE and Depth-Anything-V2 are not specifically designed for underwater conditions, future work could improve their robustness through fine-tuning or physics-based priors, thereby further enhancing the colour fidelity and geometric accuracy of our framework.

\section*{Acknowledgements}
This work was supported by the UKRI MyWorld Strength in Places Programme(SIPF00006/1) and EPSRC ECR(EP/Y002490/1).

\bibliography{egbib}
\end{document}